\documentclass[letterpaper, 10 pt, conference]{ieeeconf}  

\usepackage{graphicx}
\usepackage{sidecap}
\usepackage{siunitx}

\IEEEoverridecommandlockouts                              

\usepackage{graphics} 
\usepackage{epsfig} 
\usepackage{amsmath} 
\usepackage{amssymb}  
\usepackage[final]{changes}

\title{\LARGE \bf
Learning Probabilistic Multi-Modal Actor Models\\
for Vision-Based Robotic Grasping
}

\author{Mengyuan Yan, Adrian Li, Mrinal Kalakrishnan, Peter Pastor
\thanks{$^{1}$Mengyuan Yan is with the Department of Electrical Engineering,
        Stanford University, CA 94305, USA
        {\tt\small mengyuan@stanford.edu}. This research was conducted during Mengyuan's internship at X.}%
\thanks{$^{2}$Adrian Li, Mrinal Kalakrishnan, and Peter Pastor are with X,
        Mountain View, CA 94043, USA
        {\tt\small \{alhli, kalakris, peterpastor\}@x.team}}%
}

\begin{document}

\maketitle
\thispagestyle{empty}
\pagestyle{empty}

\begin{abstract}

Many previous works approach vision-based robotic grasping by training a value network that evaluates grasp proposals. These approaches require an optimization process at run-time to infer the best action from the value network. As a result, the inference time grows exponentially as the dimension of action space increases. We propose an alternative method, by directly training a neural density model to approximate the conditional distribution of successful grasp poses from the input images. We construct a neural network that combines Gaussian mixture and normalizing flows, which is able to represent multi-modal, complex probability distributions. We demonstrate on both simulation and real robot that the proposed actor model achieves similar performance compared to the value network using the Cross-Entropy Method (CEM) for inference, on top-down grasping with a 4 dimensional action space. Our actor model reduces the inference time by 3 times compared to the state-of-the-art CEM method. We believe that actor models will play an important role when scaling up these approaches to higher dimensional action spaces.

\end{abstract}

\section{Introduction}

Robust grasping of objects is an important capability in many robotic applications. As robots go from caged and structured industrial settings to unstructured civil environments, assumptions about the knowledge of object models and environment maps no longer hold, and learning-based methods start to show an advantage in scaling towards generalizable robotic grasping of unseen objects in unstructured environments.

Most previous methods that use deep learning for grasping formulate the problem as training a regression neural network to predict the success probability of grasp poses given RGB or depth observations~\cite{lenz2015, criticgrasping, dexnet}. Such methods require an additional component for generating grasp pose proposals which are subsequently evaluated and ranked. This process can be slow and inefficient, especially when going into high-dimensional action spaces such as full 6-DOF grasping or grasping while moving the base of the robot. In this work, we propose to directly learn the distribution of good grasp poses from self-supervised grasping trials. With recent advances in density estimation~\cite{realnvp, made, iaf}, neural network models such as {Real NVP}~\cite{realnvp} are able to approximate arbitrary distributions. Furthermore, these models can do both, efficiently generate samples from the distribution, as well as compute the probability density of given samples. We call our models actor models. With a trained actor model we are able to speed up inference by eliminating the generation-evaluation-ranking process. In addition, exploration for continuous reinforcement learning becomes more natural and adaptive compared to additive noise.

\section{Related Work}

Many prior works on end-to-end grasp prediction from visual observations formulate the problem as training a value function (critic) which estimates the probability of success given a hypothetical grasp pose~\cite{ lenz2015, criticgrasping, dexnet}. A separate component is required to generate candidate grasp poses for the critic to evaluate and rank. In~\cite{criticgrasping, qtopt}, the cross-entropy method (CEM) is used to iteratively find good actions using the value network. When the action space is high-dimensional and promising actions only occupy a small fraction of the space, such a method would require a large amount of samples or prior heuristics in order to obtain good grasp poses.

\begin{figure}[t]
    \vspace{0.2cm}
    \centering
    \includegraphics[width=0.48\textwidth]{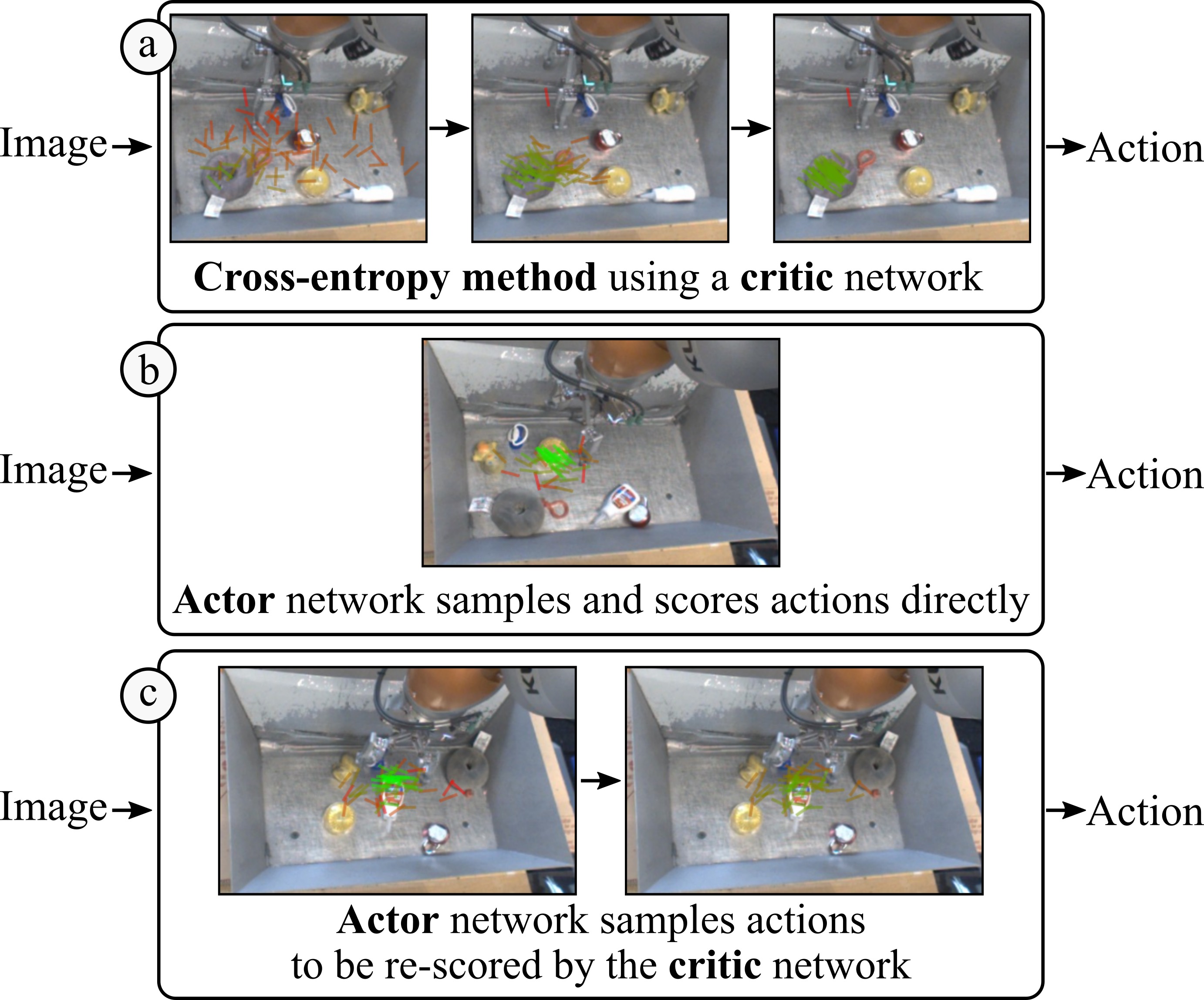}
    \vspace{-0.6cm}
    \caption{Overview diagram: The Cross-Entropy Method (a) iteratively samples from a Gaussian, scores the samples using a trained critic network, and fits a Gaussian to the elite samples, until a good grasp has been found. The actor network (b) can sample and score the actions directly in a single forward pass. The combined method (c) uses the actor to generate proposals and uses the critic network to evaluate and rank them. The actions ($\Delta x, \Delta y, \Delta z, \Theta$) are indicated with colored lines. Green lines indicate high probability/value and red lines indicate low probability/value.}
    \label{fig:kuka-network-vis}
\vspace{-0.3cm}
\end{figure}

Another approach is to directly learn a policy (actor) which predicts the optimal action given the current state. Most previous works represent this policy as either a deterministic function~\cite{gps, ddpg} or a diagonal Gaussian conditioned on the observations~\cite{soft-ac, a3c}. Such models struggle to represent multi-modal action distributions which are common in a grasping context, especially due to cluttered scenes or symmetry in objects. \added{Some works \cite{redmon15, morrison18} partially address the multi-modality problem by regressing to a grasp pose per image patch or per pixel. These approaches make assumptions on the correspondence between grasp positions in robot coordinates and in camera coordinates. This assumption makes these approaches hard to generalize to tasks beyond grasping, e.g. in-hand manipulation, where actions no longer correspond to image regions. In this work, we address the multi-modality problem by predicting a very expressive probability distribution in action space, conditioned on the whole input image, eliminating the above assumption while increasing the expressiveness of the model.} \deleted{Gaussian Mixture Models (GMM) may be employed to model more complex action distributions~\cite{mixturedensity}; however, they are still limited by their representational power, and are not very friendly to stochastic gradient descent-based optimizers.}

Generative Adversarial Networks (GAN) have received a lot of attention in modeling probability distributions for images~\cite{GAN, progressiveGAN, graspgan} and have also been used successfully in imitation learning~\cite{GAIL} settings. However, because these generators cannot compute the probability density of generated samples, a discriminator is required to provide a training signal, and balancing the interplay between these two components is difficult. Similar works in energy-based models~\cite{soft-q-learning,deep-energy} also learn generators for which the probability density of generated samples cannot be computed.

Recent work in normalizing flows~\cite{realnvp, iaf, norm-flows} makes it possible to train a neural network that can both produce samples and calculate the probability density at given points, which enables density estimation by directly maximizing the log-likelihood of observed data. We employ normalizing flows in our actor models and train them by maximizing log-likelihood of successful grasp poses. When running on the robot, the probability density of actions also serve as a confidence score for the actions.

\deleted{Our method is also related to imitation learning. In \cite{GAIL} GAN is used to match the generator model to the distribution of demonstration actions. Instead of learning from expert demonstrations, our model learns from positive examples accumulated from random trials, and continues to improve itself by executing samples from the learned distribution and receiving binary rewards in a self-supervised manner.}

\section{Motivation}

As a motivating example, we study a toy problem that captures some essential characteristics of grasping in clutter. In a $D$ dimensional action space $[0, 1]^D$, we randomly sample $10$ target points $x_1, \dots, x_{10}$ and draw hyperspheres of radius $r$ around them as regions of successful actions. We assume the value function can be learned perfectly and use $f(x)=\exp(-d(x)^2/2r^2)$ as an oracle value function, where $d(x)=\min_i\|x-x_i\|$ is the distance to the nearest target point. To evaluate the performance of CEM, we record the number of iterations required to obtain at least one successful action for which $d(x)\leq r$. At each iteration, our implementation of CEM generates $100$ normally distributed samples and uses the top $10$ elite samples to estimate the mean and variance for the next iteration.

\begin{figure}
    \centering
    \includegraphics[scale=0.29]{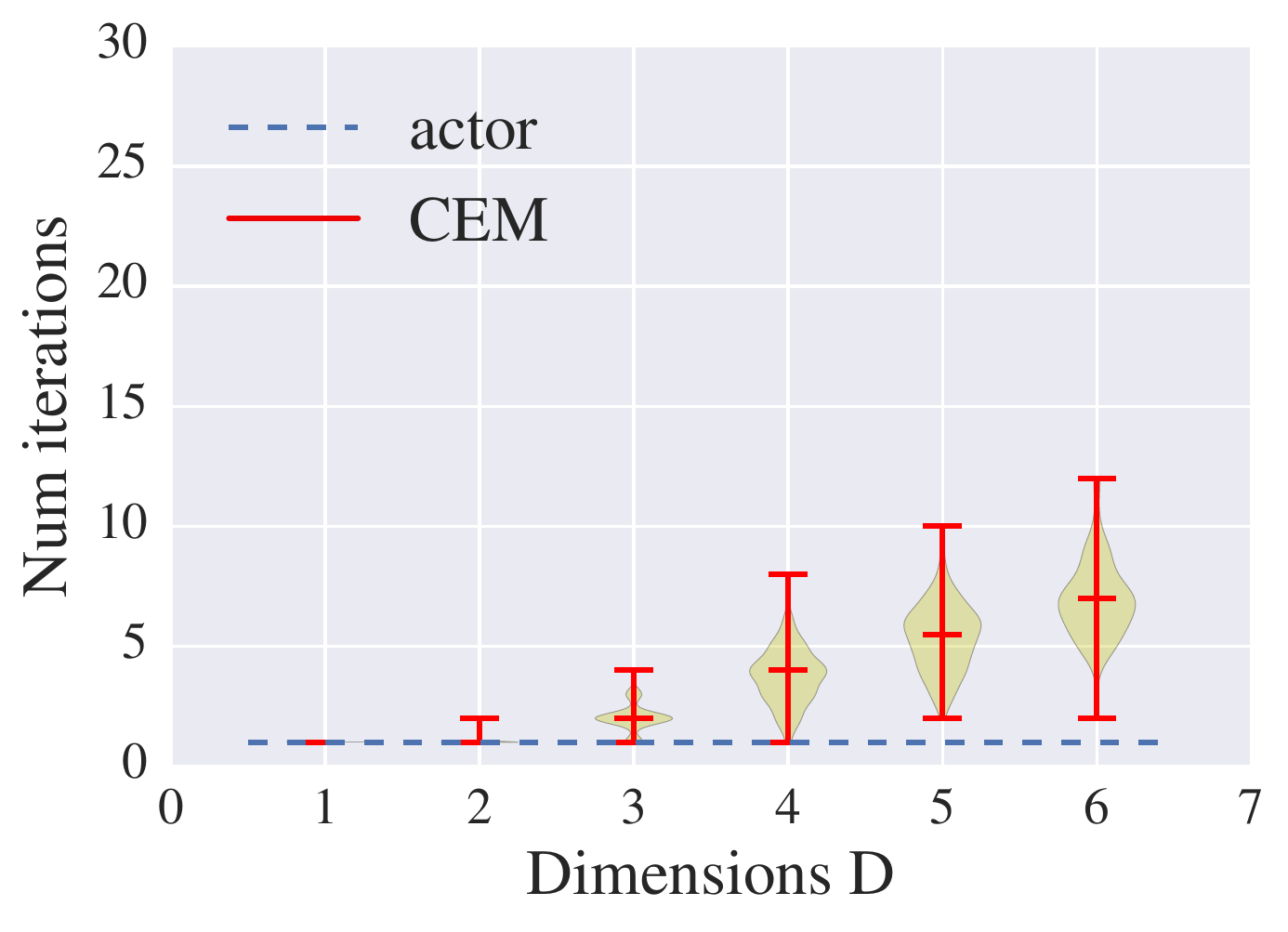}
    \includegraphics[scale=0.29]{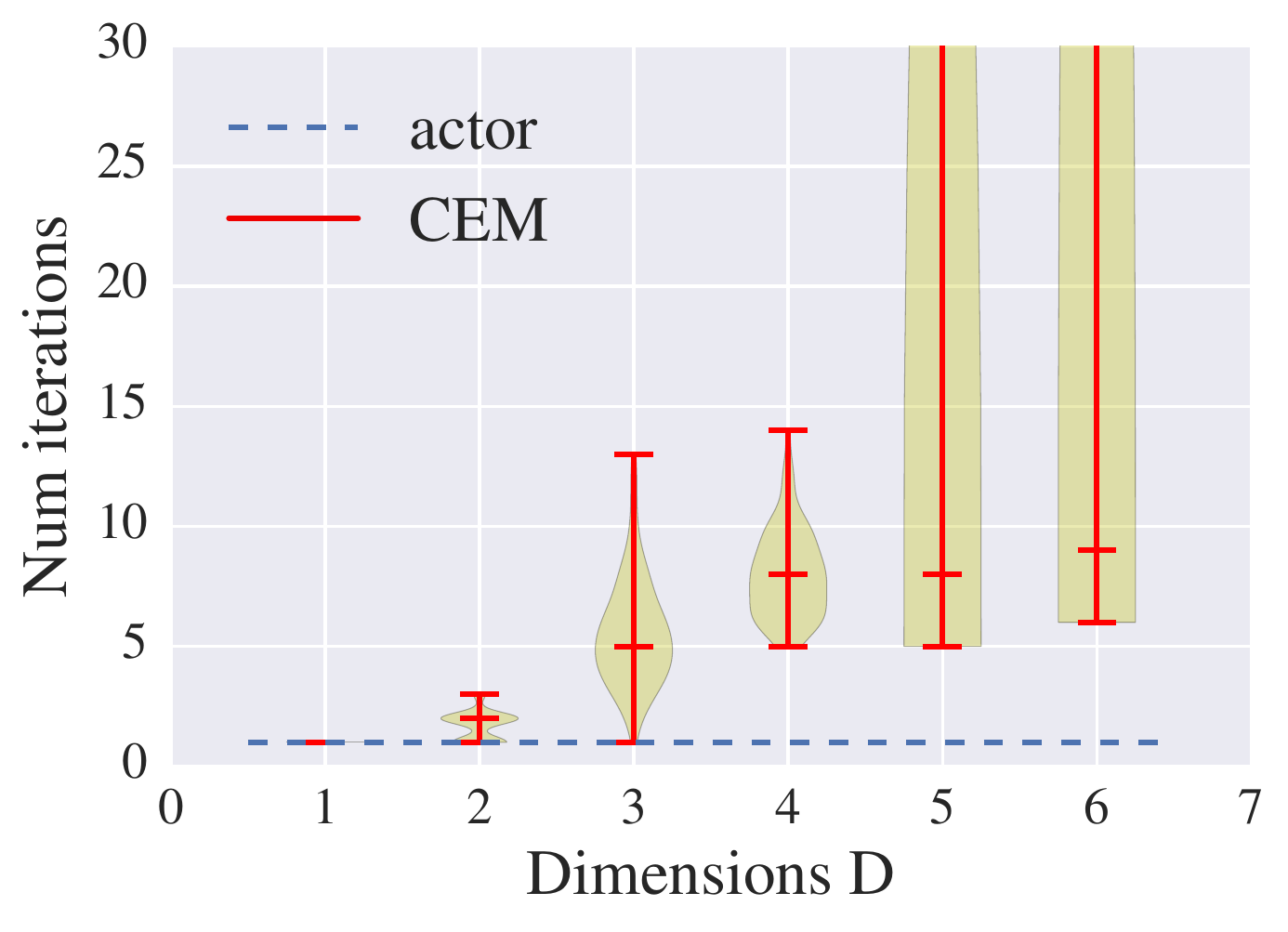}
    \vspace{-0.3cm}
    \caption{Violin plot of the number of CEM iterations required to get one successful action. Red horizontal bars represent the min, median, and max of the distribution. Left: $r=0.1$. Right: $r=0.03$. When $r=0.03$ some experiments cannot find a successful action within 50 iterations. Blue dashed lines represent the actor model, which once trained, can always predict at least one successful action with only 1 iteration.}
    \label{fig:motivation}
    \vspace{-0.3cm}
\end{figure}

We repeat this experiment $100$ times for different settings of the hyperparameters $D$ and $r$. The distributions are plotted in Fig.~\ref{fig:motivation} for $D=1, \dots, 6$ and $r=0.1, 0.03$. The average number of iterations required to find a successful action grows exponentially with the dimension of the action space, and also grows quickly when raising the required precision of the task. In some experiments, CEM failed to even find a successful action within a maximum of $50$ iterations.

These experiments illustrate the drawbacks of using CEM to select optimal actions at inference time. For higher-dimensional action spaces and high-precision tasks, iteratively optimizing on a multi-modal value function is inefficient. This is especially undesirable when action selection needs to happen inside a high-frequency control loop, such as in robotics.

On the other hand, we can train an actor model to predict the distribution of successful actions given $x_1, \dots, x_{10}$ as the input state. With enough training, the actor can always predict a successful action with a single iteration, by generating $100$ samples from the predicted distribution and selecting the one with highest probability density. As the dimension of the action space increases, the state space also expands exponentially, thus requiring more training time; however, the inference time remains constant.

\section{Method}
With some mild assumptions, e.g. the environment has no obstacles to avoid, and objects are not too densely packed so that pre-grasp manipulation is necessary, learning to grasp can be seen as predicting the grasp poses that lead to high success probability. We approach this problem by training a neural density model that approximates the ground truth conditional distribution of successful grasp poses.

\subsection{Neural density models}
There are several types of neural network models that are very powerful in representing probability distributions. In this work we study the Gaussian mixture model \cite{mixturedensity} and {Real NVP} model \cite{realnvp}, and a combination of both, which we call the mixture of flows (MoF) model. We briefly describe the three models below.

\paragraph{Gaussian mixture model}
A neural network is trained to predict the centers \added{$\mu_i$}, variances \added{$\sigma_i$}, and weights \added{$w_i$} of $k$ multivariate diagonal Gaussians, where $k$ is determined according to the task. For predicting weights \added{$w_i$}, a softmax layer is used as the last layer to satisfy the constraint that $\sum_i w_i = 1$. GMM is the simplest probabilistic model to approximate a multi-modal distribution. However, GMM is not friendly to stochastic gradient optimization. With maximum log-likelihood as the objective, there are saddle points in the optimization landscape that are hard to escape from, even for momentum or Adam optimizers. As a simple example, assume the ground truth distribution is uniform on $[-0.1, 0.1] \cap [0.9, 1.1]$, \added{and we are approximating this distribution with GMM $k=2$}. It is easy to verify that there is a saddle point at $\mu_1=\mu_2=0.5$, $\sigma_1=\sigma_2=0.5$, $w_1=w_2=0.5$\deleted{, where $\mu_i, \sigma_i, w_i$ are the centers, standard deviations and weights of the GMM with $k=2$}. In experiments we observed that the optimization got stuck on such saddle points very often, unless the parameters of the GMM were carefully initialized, which requires prior knowledge of the task at hand.

\paragraph{Real-valued non-volume preserving ({Real NVP}) transformations}
{Real NVP} transformations are bijective mappings between the latent space and the prediction space. If the probability distribution in the latent space is known, then the distribution in the prediction space can be calculated as 
$$
p(x) = p(z)\left| \det\left(\frac{\partial{x}}{\partial{z}}\right) \right|^{-1},
$$
where $x$ is a point in the prediction space and $z$ is the corresponding point in the latent space, and $z$ is calculated from $x$ using the inverse function. A multivariate normal distribution $\mathcal{N}(0, I)$ is used for the prior distribution $p(z)$.

For a general fully-connected neural network, it is time-consuming to compute the determinant of the partial derivative matrix $\frac{\partial{x}}{\partial{z}}$, and its derivative with respect to network parameters. The network is also not guaranteed to be a bijective function. In \cite{realnvp} the authors proposed a special way of constructing the neural network to solve these problems. The latent space is split into two orthogonal sub-spaces, $z_{1:d}$ and $z_{d+1:D}$, and the transformation \added{$x=f(z)$} is defined to be the composition of a series of affine \replaced{transformations $f(z)=g_n(...(g_2(g_1(z))))$}{coupling layers}. \replaced{Each affine transformation (also called coupling layers) $g_i$ has}{of} the form
\begin{eqnarray*}
y'_{1:d} &=& y_{1:d}, \\
y'_{d+1:D} &=& y_{d+1:D}\odot\exp(s\added{_i}(y_{1:d}))+t\added{_i}(y_{1:d}),
\end{eqnarray*}
or similarly
\begin{eqnarray*}
y'_{d+1:D} &=& y_{d+1:D}, \\
y'_{1:d} &=& y_{1:d}\odot\exp(s\added{_i}(y_{d+1:D}))+t\added{_i}(y_{d+1:D}),
\end{eqnarray*}
where the functions $s\added{_i}(y)$ and $t\added{_i}(y)$ are neural networks that predict the vectors of log-scale and translation of the affine transformation, and $\odot$ is the Hadamard (or element-wise) product. The neural networks \added{$s_i$ and $t_i$} may optionally be conditioned on features of the input \replaced{observations}{state}.

By alternating between the two coupling layers, the composed transformation can be arbitrarily complex. This class of transformations has two desirable properties: their inverse function can be easily computed by inverting each affine coupling layer, and the determinant of partial derivative $\frac{\partial{y'}}{\partial{y}}$ for each layer can be easily calculated as $\prod \exp(s(y))$. As a result, we are able to efficiently sample from the predicted distribution, and also compute the probability of given data under the predicted distribution.

\paragraph{Mixture of flows model}
It is straight forward to use a multivariate normal distribution in the latent space for {Real NVP} models. However, in experiments we observe that it is difficult for {Real NVP} models to learn a cluster-like distribution, where the support of the target distribution is separated into modes, instead of a continuous region. To make the model more expressive, we combine Gaussian mixture and {Real NVP} into a mixture of flows (MoF) model, where the latent space distribution is a learnable Gaussian mixture, and each Gaussian in the latent space is transformed by an independent {Real NVP} transformation. The MoF model combines the good from both worlds. It does not suffer from the saddle point problem of GMM, and the model can easily use different Gaussian components to model different modes in the action space.

\subsection{Actor model training}

Using neural density models enables us to directly train an actor model by maximizing the log-likelihood, as opposed to GAN-style adversarial training where a critic or discriminator is required. With a dataset of successful grasps $\mathcal{D}$, the training loss is
$$
\mathcal{L} = - \mathbb{E}_{s,a \in \mathcal{D}} \{\log(\pi(a|s))\}.
$$

If we give binary reward of $r=1$ to successful grasps and $r=0$ to failed grasps, this loss is equivalent to
$$
\mathcal{L} = - \mathbb{E}_{a \sim \pi_B(a|s)} \{r(s,a)\log(\pi(a|s))\}.
$$
where $\pi_B(a|s)$ is the behavior policy used to collect the dataset. Our training loss is equivalent to minimizing the KL divergence $D_{KL}(\pi_B(a|s)r(s,a) \| \pi(a|s))$.
When the behavior policy is uniform random across the action space, assuming our density model is able to approximate arbitrary probability distributions, the optimal policy is $\pi(a|s) \propto r(s,a)$, and covers every successful action. 

As the task gets more difficult, the success rate of a random policy can be low, and collecting a dataset of successful grasps from random trials can be inefficient. We can also sample actions from the actor model instead of a random distribution to add data into the dataset. Inferring actions from the action model increases grasp success rate and makes learning more efficient. In this case, the training loss needs to be adapted. Since $\pi_B(a|s) \rightarrow \pi(a|s)$, minimizing the KL divergence between the unnormalized distribution $\pi(a|s)r(s,a)$ and $\pi(a|s)$ is prone to mode missing. Maximum entropy regularizer is added to the training loss to prevent mode missing. The loss becomes
\begin{equation*}
\resizebox{.48 \textwidth}{!} 
{
$
\mathcal{L} = - \mathbb{E}_{a \sim \pi_B(a|s)}\{r(s,a)\log(\pi(a|s))\} + \alpha \mathbb{E}_{a \sim \pi(a|s)}\log(\pi(a|s)),
$
}
\end{equation*}
where $\alpha$ is the relative weight between the two losses. It is not hard to prove that the action distribution will converge to $\pi(a|s) \propto \exp(r(s,a)/\alpha)$.

\section{Robot grasping overview}

We demonstrate the result of training actor models for vision-based robotic grasping. An illustration of our grasping setup is shown in Fig. \ref{fig:kuka-setup}. To demonstrate the advantage of a probabilistic actor instead of a deterministic one, all our experiments have multiple objects in the workspace, thus the distribution of good grasps is multi-modal.

The observation sent to the actor model includes the robot’s current camera observation, a $472\times472$ RGB image, recorded from an over-the-shoulder monocular camera (see Fig. \ref{fig:kuka-setup}), and an initial image taken before the arm is in the scene. The action $a$ is a 4 dimensional top-down grasp pose, with a vector in Cartesian space $t \in \mathbb{R}^3$ indicating the desired change in the gripper position, and a change in azimuthal angle encoded via a sine-cosine encoding $r \in \mathbb{R}^2$. The gripper is scripted to go to the bottom of the tray and close on the final time step.

In simulation, grasp success is determined by raising the gripper to a fixed height, and checking the objects' poses. For the real robots, the post-grasp and the post-drop images are subtracted, both without the arm in the view. Only if the two images are significantly different, because an object was dropped back into the tray, a grasp is determined as successful. This labeling process is fully automated to achieve self-supervision.

The neural network for the actor model consists of 7 convolution layers to process the image, followed by a spatial softmax layer to extract 128 feature points. The coordinates of the feature points are then processed with 2 fully connected layers to produce the final representation of the input images, which is used to predict the parameters of the Gaussian mixture, and/or concatenated with the latent code to predict the log scale and translation for {Real NVP's} affine coupling layers.

Although the actor model is trained to predict good grasp poses in one step, our robots take multiple actions for each grasp trial, both for data collection and for evaluation. For data collection, the number of actions taken is random between $3$ and $10$. To transform the recorded grasp trials into data samples suitable for training the actor model, at each step the action is determined by the difference between the final grasping pose and the current gripper pose, and grasp success determined at the end of the trial is used for every step in the process. For evaluation, \replaced{the robot will close its gripper and end one grasp trial if it has converged to a grasp pose, or a maximum of 10 actions is reached. Experimentally we define convergence as if the selected action is within $5$mm movement in Cartesian space and $2^{\circ}$ rotation for the actor, and if the predicted value for zero action is above $0.95$ of the highest sample's value for the critic.}{the maximum number of actions taken has been fixed to $10$.} 

In simulation, we evaluate the performance of our actor model with pure off-policy data as well as on-policy data. When training with only off-policy data, the robots moves randomly within the workspace, and successful grasps are extracted. When training on-policy, the initial $100k$ successful transitions are collected by random policy, after which the actor model is used to sample actions, and successful grasps are added to the data buffer. We use $1000$ simulated robots and $3$ GPUs to collect data and perform training asynchronously. 

We also evaluated our method on real KUKA robots. In this case our models are trained with a dataset of grasps previously collected and used in \cite{graspgan}, which has 9.4 million training samples in total, including 3.6 million successful samples.

\section{Experiments}

\subsection{Representation power of neural density models}
We demonstrate the advantage of our MoF model compared to other neural density models on a toy task as well as on robotic grasping. 
For the toy task, we randomly sample 5 points $p_0, p_1, \dots, p_4$ in a $10\times10$ square. The coordinates of the 5 points are observations, and the distribution of good actions is defined as a mixture of Gaussians centered at $p_1-p_0, \dots, p_4-p_0$ with standard deviation $0.5$. The GMM model has 1 linear layer and 4 components, the {Real NVP} model has 4 affine coupling layers and each translation and log scale function has 2 fully connected layers. For the MoF model, the base distribution in the latent space uses the same model as GMM, and each {Real NVP} branch has the same architecture as the {Real NVP} model.

\begin{figure}[thpb]
  \centering
  \includegraphics[scale=0.25]{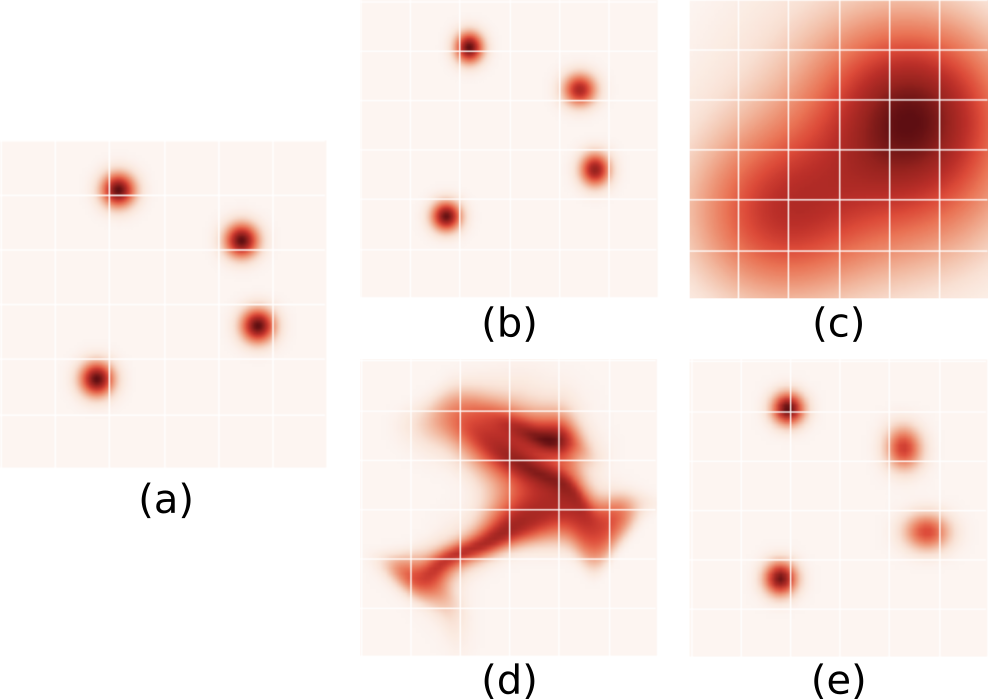}
  \vspace{-0.2cm}
  \caption{Probability density predictions on the toy task. (a) Ground truth distribution. (b) GMM can be trained to match the ground truth with initial standard deviation $1.0$. (c) When initial standard deviation is $10.0$, GMM failed to converge to the ground truth. (d) {Real NVP} model predicts a continuous region of high probability and can not match the ground truth well. (e) MoF model prediction matches the ground truth well, and is the same for initial standard deviation $1.0$ or $10.0$.}
  \label{fig:toy-modelcompare}
\end{figure}

Visualizations of each model's prediction is shown in Fig.~\ref{fig:toy-modelcompare}. Due to the saddle point problem in GMM models, its convergence is dependent on the initial value of Gaussian variances. The {Real NVP} model prediction covers all 4 Gaussians in the ground truth distribution, but also has a significant probability mass in areas that are not supported by the ground truth. The MoF model can represent the ground truth distribution well, and is also robust to changes in the initial variances of the base Gaussian mixture.

\begin{figure}
    \centering
    \includegraphics[width=0.46\textwidth]{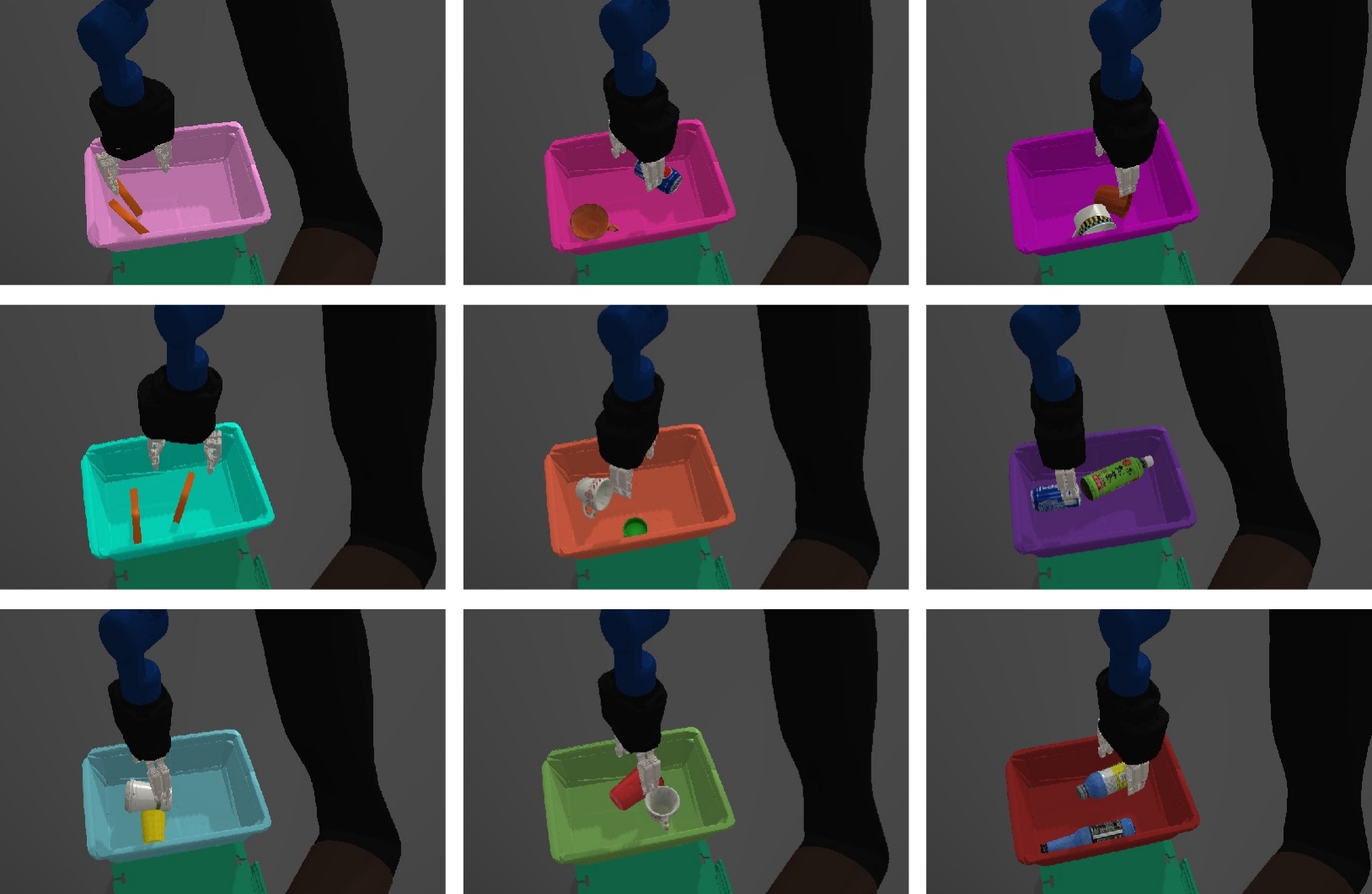}
    \vspace{-0.2cm}
    \caption{Cropped visualization of camera images in simulation. Two images on the top left shows grasping of blocks and other images show sample objects from the bottles and cups set.}
    \label{fig:sim_dv}
    \vspace{-0.2cm}
\end{figure}

\begin{figure}[thpb]
  \centering
  \vspace{-0.2cm}
  \includegraphics[width=0.3\textwidth]{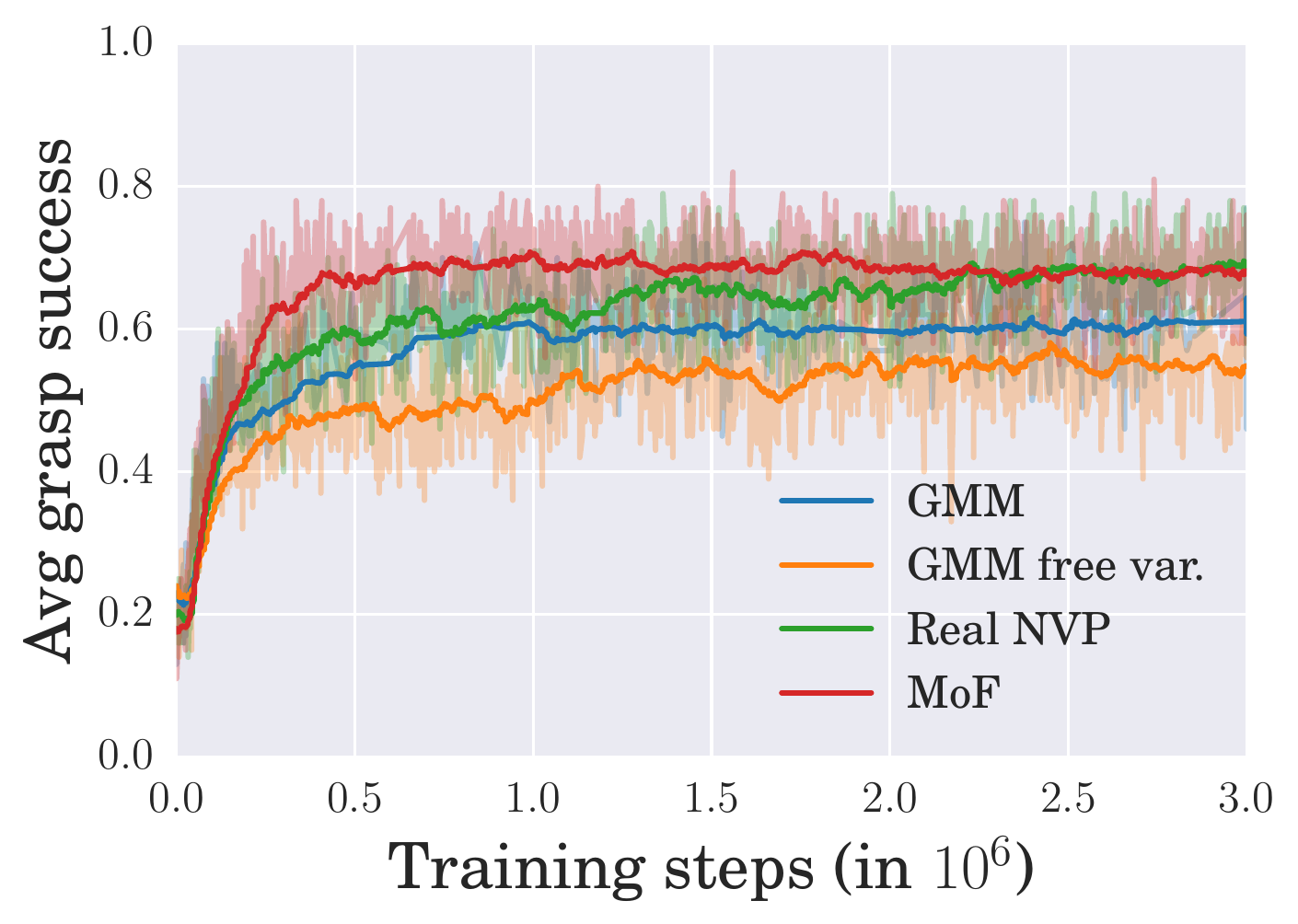}
  \vspace{-0.3cm}
  \caption{Grasp success rate during training with GMM using trainable variance (orange), GMM using fixed variance (blue), Real NVP (green), and MoF model (red). The GMM using trainable variance is less stable and performs worse than the GMM using a fixed variance. The MoF model achieves the highest grasp success rate.}
  \label{fig:grasping-modelcompare}
  \vspace{-0.4cm}
\end{figure}

We also evaluate the representation power of the models on robotic grasping. The models are trained on an off-policy dataset of 1.8M successful grasps. Training and test objects are a set of 30 drink bottles and cups. Some examples of the objects are shown in Fig.~\ref{fig:sim_dv}. The grasp success rates are plotted in Fig.~\ref{fig:grasping-modelcompare}. Our MoF model achieves the highest grasp success rate. For the GMM model, trainable variance is less stable and performs worse than a fixed variance determined beforehand according to our knowledge of the task. 

\subsection{Data efficiency and inference speed of actor vs. critic models}
We compared our method with training a critic model \cite{criticgrasping}, where a cross-entropy method (CEM) optimization process is used to find good actions during evaluation. 
We collected datasets of different sizes by running a random policy. For training the critic model, both successful and failed grasps are used, while for training the actor model only successful grasps are used. However, we report dataset size as the number of actions tried, including successful and failed ones, even for the actor model. This corresponds to the time required to collect training data, although actor models are at a disadvantage in this comparison due to low success rate of random trials. We evaluate both the actor model and the critic model on 2 grasping tasks with different objects. In the first task, there are 2 blocks in the basket that have the same appearance, and in the second task the basket has two objects from a set of 30 different commodity drink bottles and cups. Camera images from the robot for both tasks are shown in Fig. \ref{fig:sim_dv}. Random trial success rate is $9\%$ on the blocks and $6\%$ on the bottles and cups.  

During evaluation, the actor model samples $64$ actions for each input observation and picks the action that is kinematically feasible and has the highest probability. For the critic model, the CEM process runs for 3 iterations with population $64$ and picks $10\%$ top samples for estimating the mean and variance for the next iteration, the sample with the highest score is executed on the last iteration.  

\begin{figure}[t]
  \centering
  \includegraphics[scale=0.27]{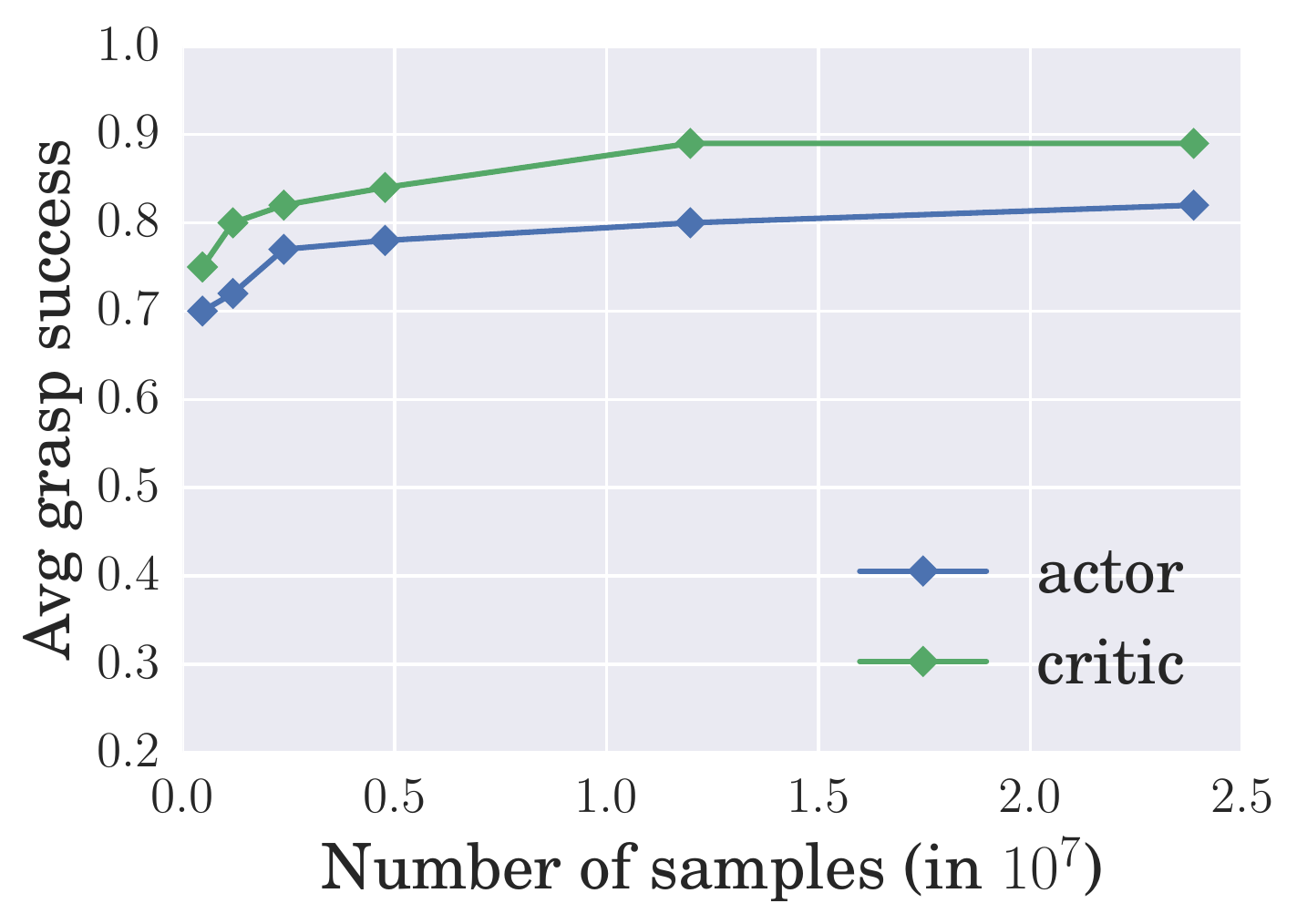}
  \includegraphics[scale=0.27]{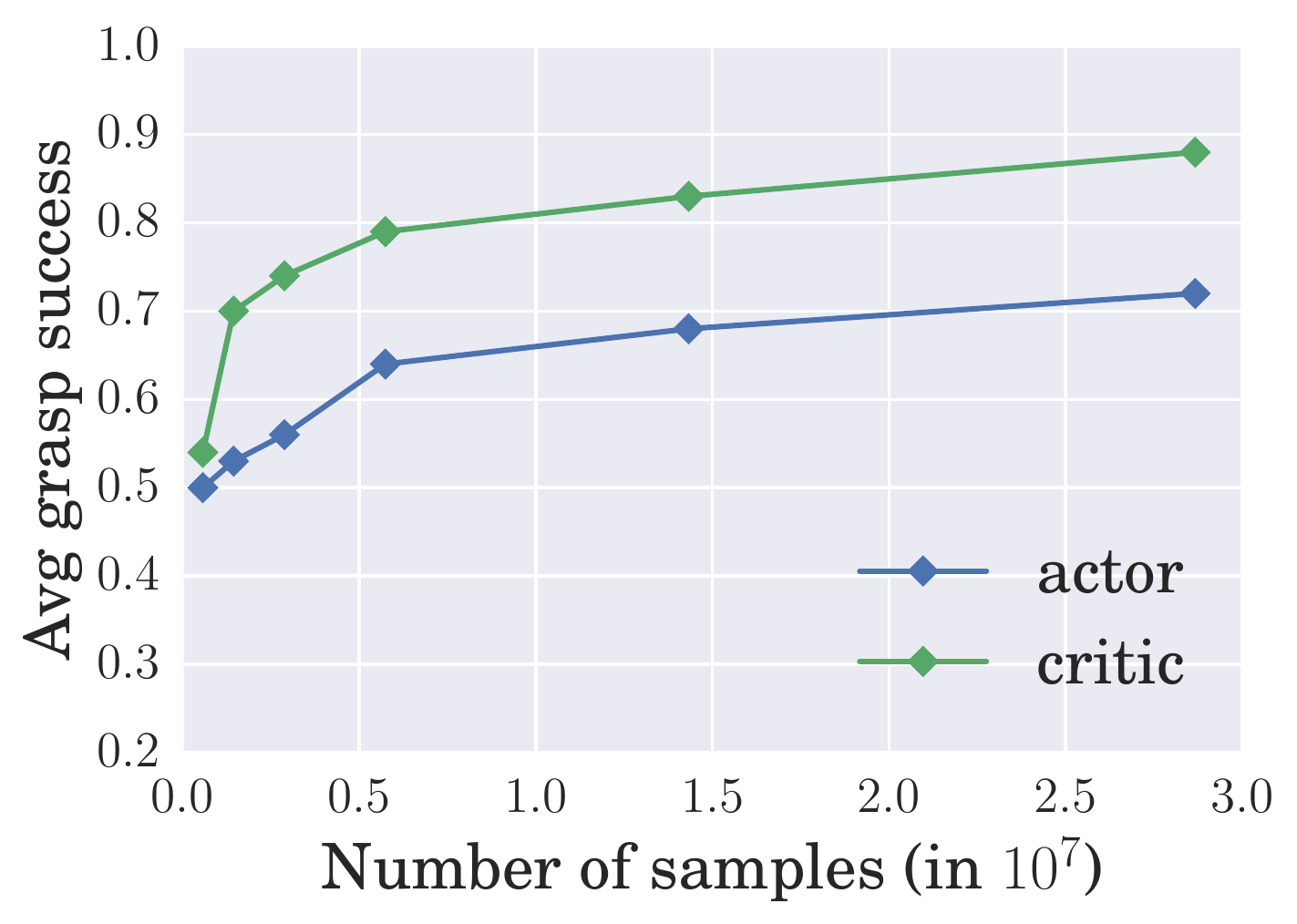}
  \vspace{-0.3cm}
  \caption{Data efficiency of actor and critic models on grasping blocks (left) and grasping of bottles and cups (right). The trend of growth is similar for both the actor and the critic models, although the success rate for the actor model is lower than the critic model.}
  \label{fig:offpolicy}
  \vspace{-0.3cm}
\end{figure}
  
Figure~\ref{fig:offpolicy} plots the grasp success rate in evaluation as the number of training examples increases. The trend of growth is similar for both the actor and the critic models, although the success rate for the actor model is lower than the critic model, especially on the more difficult set of objects. Our hypothesis is that it is easier for the network to judge if a hypothetical action will be successful, by giving attention to only the area around the destination, while the actor model has to digest the whole image and learn all modes of possible successful actions. 


Because the CEM optimization takes 3 iterations, the inference time for the CEM policy is 3 times longer than the inference time of the actor policy. The CEM policy inference also takes significantly more memory on the GPU, because the visual features need to be replicated to merge with the feature of each grasp pose proposal, while the actor can predict the parameters for the probability distribution once, and sample many grasp poses with very little overhead.

\subsection{Actor models as natural exploration policy}
\begin{figure}[b]
  \centering
  \includegraphics[width=0.23\textwidth]{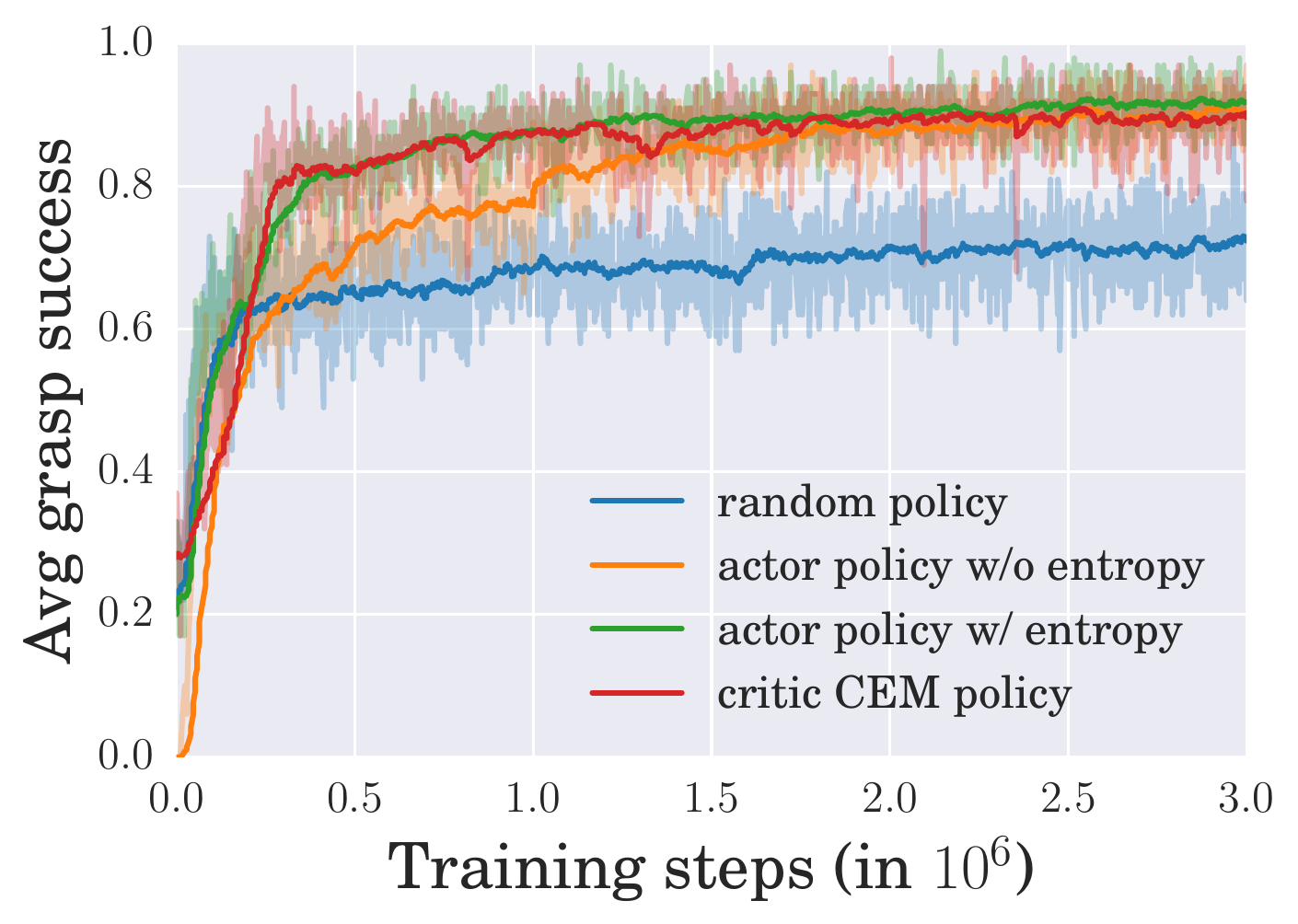}
  \includegraphics[width=0.23\textwidth]{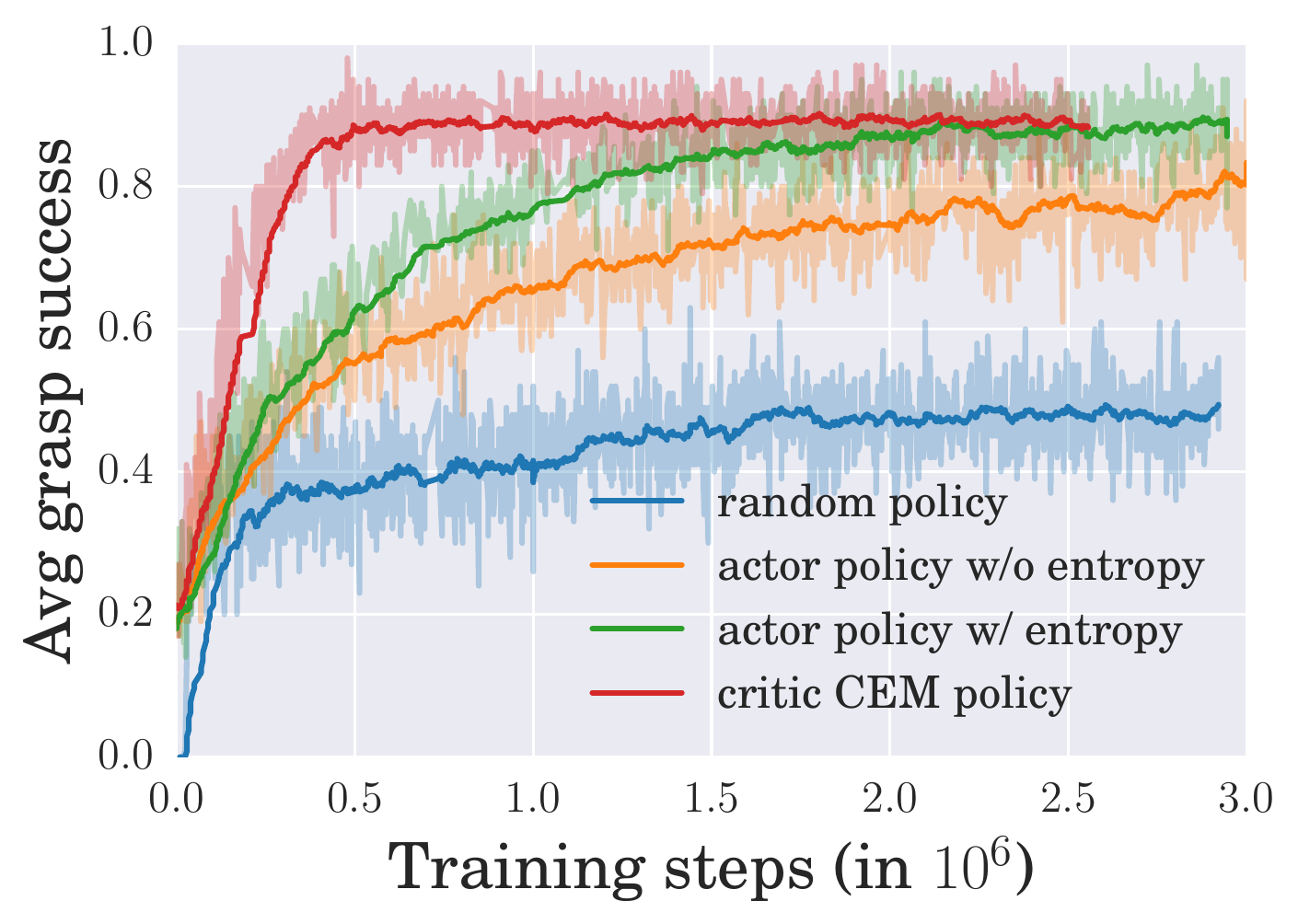}
  \vspace{-0.3cm}
  \caption{On-policy training of actor and critic models on grasping of blocks (left) and grasping of bottles and cups (right). By collecting data using samples from the actor model (green and orange), the model can learn significantly faster than if data is collected with random actions (blue). Adding entropy regularizer in the training loss (green) also improves training speed and performance. The actor model can achieve the same final performance as the critic model (red), although for the harder task the critic model improves faster at the beginning of the training.}
  \label{fig:onpolicy}
\end{figure}
Actor models provide a natural way of exploration for on-policy training. Once the actor model is trained with a small amount of off-policy data, it can be used to sample actions for collecting more grasping data, with a significantly higher rate of success.

We compare training the actor model using a replay buffer~\cite{mnih2015}, where data is collected by sampling actions randomly, versus sampling from the actor model. We also compare with training a critic model on-policy, where data is collected by running CEM with the critic model. 

Grasp success rates are plotted in Fig.~\ref{fig:onpolicy} as the training progress. Using actor model to sample actions and collect data has a clear advantage compared to using random actions. This advantage is more obvious when the task requires more precision and the random policy gets fewer successes. It is also clear that having an entropy regularizer in the training loss helps to improve training speed and final performance of on-policy training. Finally, the actor and the critic can achieve the same final success rate, although for the harder task of grasping bottles and cups, the critic improves faster at the beginning of training. \added{In both tasks the highest grasp success rate reaches about 90\%, where most failures are due to objects being too close to the corners of the bin and the gripper collides with the bin before grasping.}

\subsection{Real robot experiments}
We trained and evaluated the actor model and the critic model on real KUKA robots. Both models are trained on the same dataset of real robot grasps, but for the actor model only successful grasps are used. We evaluate the actor model by predicting 64 samples at each time step and taking the action with the highest probability density. \deleted{The actor will close the gripper and terminate one grasp if the selected action is within $5\si{mm}$ movement in Cartesian space and $2^{\circ}$ rotation, or if a maximum of 10 time steps is reached.} For evaluating the CEM method using the critic model, we set the initial Gaussian to have a standard deviation of $15\si{cm}$ in horizontal direction, $6\si{cm}$ in vertical direction and $90^{\circ}$ in rotation. This distribution is chosen to cover the space of the tray. The CEM is run for 3 iterations (see \cite{criticgrasping}) and the action with highest predicted value is selected. We also evaluated a policy that combines both the actor model and the critic model, where the actor model predicts 64 samples, and the samples are evaluated by the critic model. Finally the action scored highest by the critic model is selected. 

\begin{figure}[t]
    \centering
    \includegraphics[width=0.48\textwidth]{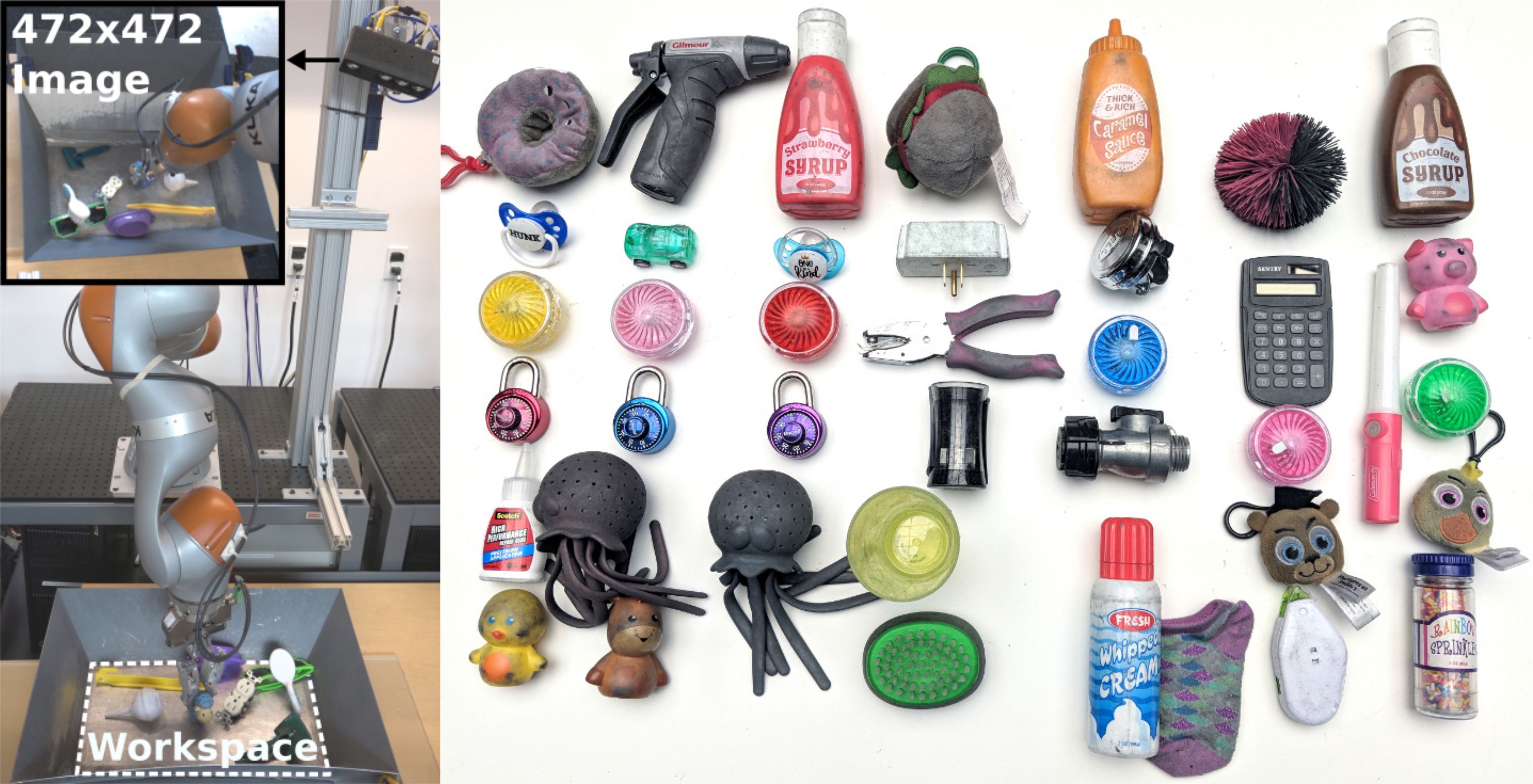}
    \vspace{-0.6cm}
    \caption{The experimental setup on the left and the set of test objects on the right. The test objects have not been seen during training.}
    \label{fig:kuka-setup}
    \vspace{-0.2cm}
\end{figure}

The experiment consisted of 6 sets of 30 grasp attempts on each of the 7 KUKA robots, totalling $6\times 30\times 7=1260$ grasp attempts. Each robot was presented with 5-6 objects from the test set, shown in Fig.~\ref{fig:kuka-setup}. The average success rate of each run using the three presented methods are summarized in the table below.

{
\parindent0pt
\begin{table}[h!]
\begin{center}
\scalebox{0.86} {
\begin{tabular}{| l | c | c | c | c | c | c | c |}
\hline
{Method} & Run 1 & Run 2 & Run 3 & Run 4 & Run 5 & Run 6 & Avg. \\ \hline
{\bf CEM} & 78.1\% & 76.2\% & 79.0\% & 76.1\% & 82.3\% & 80.5\% & {\bf 78.7\%} \\ \hline
{\bf Actor only} & 80.0\% & 80.0\% & 70.4\% & 77.1\% & 77.1\% & 75.7\% & {\bf 76.7\%} \\ \hline
{\bf Actor+Critic} & 86.9\% & 81.4\% & 82.8\% & 80.0\% & 81.4\% & 84.7\% & {\bf 82.8\%} \\ \hline
\end{tabular}
}
\label{tab:results}
\end{center}
\end{table}
\vspace{-0.5cm}
}

\begin{figure}[t]
    \centering
    \includegraphics[width=0.47\textwidth]{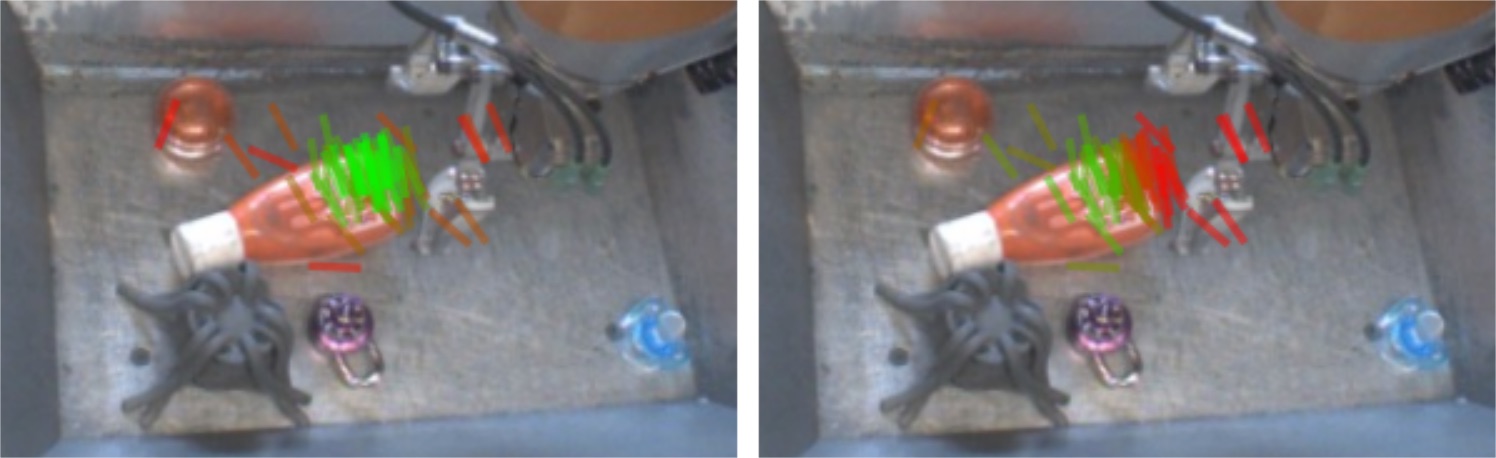}
    \vspace{-0.3cm}
    \caption{Cropped view as seen by the over-the-shoulder camera. The samples generated by the actor (colored lines in the left image) are re-scored by the critic (colored lines in the right image). Green lines indicate higher probability of success, red lines indicate unlikely grasps. The actor is able to predict good samples in one shot and the critic favors robust grasps.}
    \label{fig:kcam_image}
    \vspace{-0.4cm}
\end{figure}

Visualization of the actor model and the critic model predictions are shown in Fig. \ref{fig:kuka-network-vis}. The critic model predicts a smooth function over the workspace and the CEM samples gradually concentrate towards the high valued region that covers one of the objects. The actor model directly predicts promising samples that usually concentrates on the object closest to the gripper. Occasionally, the actor model predicts samples on the boundary of objects that would result in unstable grasps, see e.g. Fig.~\ref{fig:kcam_image}. The critic model predicts higher values for samples that are closer to the object center and thus more stable, and help improve grasp success.

\section{Conclusions and discussions}
We proposed an alternative way of vision-based robotic grasping. Instead of training a critic model that evaluates grasp proposals, we directly train a neural density model to approximate the conditional distribution of successful grasp poses given input images. We demonstrated on both simulation and real robot that the proposed actor model achieves similar performance compared to the critic model using CEM for inference. On the top-down grasping task with 4 dimensional action space, our actor model reduces inference time by 3 times compared to the state-of-the-art CEM method, at the cost of longer training time. Going into higher dimensional action space, we believe actor models will be more promising and scalable, while the CEM method will take exponentially longer inference time or even fail to find good solutions from the critic model.

Our proposed actor model also has limitations. One of the limitations is that our model only uses successful grasps as training data. While the density model normalizes over the action space, thus assumes every action that is not included in the dataset of successful grasps is failure, there may be additional information in the failed examples that can be helpful for improving the actor model. As future work, our actor model can be trained jointly with a critic model, where the binary reward from the dataset can be replaced by the value predicted by the critic model. Another way to incorporate information from failed trials is to train two separate actor models, one that predicts the distribution of successful actions (success actor model), and one that predicts the distribution of all actions tried (prior actor model). During evaluation we can predict samples from the success actor model, and evaluate the samples by taking the quotient of probability densities of the success actor model and prior actor model.

\addtolength{\textheight}{-12cm}   






\bibliographystyle{plain}
\bibliography{reference.bib}

\end{document}